%% file: main.tex
\documentclass[10pt, journal]{IEEEtran}        
\usepackage{orcidlink}
\usepackage[style=ieee, maxcitenames=1, mincitenames=1]{biblatex} 

\bibliography{bibliography.bib} 

\usepackage{amsmath,amssymb,amsfonts}
\usepackage{graphicx}
\usepackage{pifont}
\usepackage{cleveref}

\usepackage{amsthm}
\usepackage{xcolor}
\usepackage[figuresright]{rotating}

\usepackage{graphicx}
\graphicspath{{/}{fig/}}

\usepackage{array}
\usepackage{textcomp}
\usepackage{xcolor}
\usepackage{multirow}
\usepackage{booktabs}

\usepackage{mathtools}
\usepackage{breqn}
\usepackage{float}

\usepackage{pgfplots}
\pgfplotsset{compat=1.7}
\usepgfplotslibrary{groupplots}

\usepackage{multirow,tabularx}
\usepackage{hyperref}
\usepackage{algorithmic}
\usepackage[vlined, ruled, shortend]{algorithm2e}
\usepackage{tikz}
\newcommand{\tikzcircle}[2][red,fill=red]{\tikz[baseline=-0.5ex]\draw[#1,radius=#2] (0,0) circle ;}%
\usepackage{placeins}
\usepackage{enumerate}

\usepackage{bbding}
\usepackage{etoolbox}
\makeatletter
\patchcmd{\@makecaption}
  {\scshape}
  {}
  {}
  {}
\makeatletter
\patchcmd{\@makecaption}
  {\\}
  {.\ }
  {}
  {}
\makeatother
\DeclareMathOperator{\arctantwo}{arctan2}

\newlength\figureheight
\newlength\figurewidth
\title{
    Shared Control of Holonomic Wheelchairs through Reinforcement Learning \\
}

\author{
    Jannis Bähler\IEEEauthorrefmark{2},
         Diego Paez-Granados*\IEEEauthorrefmark{2}\IEEEauthorrefmark{3},
         Jorge Peña-Queralta*\IEEEauthorrefmark{2}\IEEEauthorrefmark{4} \\[+1em]

         \IEEEauthorrefmark{3}\href{https://www.paraplegie.ch/spf/en/}{Swiss Paraplegic Research, SPF, Nottwil} \\[+.42em]
         \IEEEauthorrefmark{2}\href{https://scai.ethz.ch/}{SCAI Lab, D-HEST, Swiss Federal School of Technology in Zurich - ETH Zurich} \\[+.42em]
         \IEEEauthorrefmark{4}\href{https://www.zhaw.ch/en/engineering/institutes-centres/cai}{Centre for Artificial Ingelligece, Zurich University of Applied Sciences - ZHAW. Switzerland}.
    \thanks{*Equal advising and project management.}
}

\begin{document}

\maketitle
\thispagestyle{empty}
\pagestyle{empty}

\input{sec/00_Abstract.tex}
\IEEEpeerreviewmaketitle

\input{sec/01_Intro}

\input{sec/02_RelatedWorks}

\input{sec/03_Background}
\input{sec/04_Methodology}
\input{sec/05_Experiments}
\input{sec/06_Conclusion}

\printbibliography

\textcolor{white}{\[EOD\]}
\newpage
\appendices
\input{sec/XX_Appendix}

\end{document}

%% file: sec/00_Abstract.tex

\begin{abstract}%
    \label{sec:abstract}%
    Smart electric wheelchairs can improve user experience by supporting the driver with shared control. State-of-the-art work showed the potential of shared control in improving safety in navigation for non-holonomic robots. However, for holonomic systems, current approaches often lead to unintuitive behavior for the user and fail to utilize the full potential of omnidirectional driving. Therefore, we propose a reinforcement learning-based method, which takes a 2D user input and outputs a 3D motion while ensuring user comfort and reducing cognitive load on the driver.  Our approach is trained in Isaac Gym and tested in simulation in Gazebo. We compare different RL agent architectures and reward functions based on metrics considering cognitive load and user comfort. We show that our method ensures collision-free navigation while smartly orienting the wheelchair and showing better or competitive smoothness compared to a previous non-learning-based method. 
    We further perform a sim-to-real transfer and demonstrate, to the best of our knowledge, the first real-world implementation of RL-based shared control for an omnidirectional mobility platform.

\end{abstract}


%% file: sec/01_Intro.tex

\section{Introduction}
\label{sec:intro}

\IEEEPARstart{P}{owered} wheelchairs can significantly enhance mobility and autonomy for people with various physical limitations (e.g. spinal cord injuries). This work considers smart wheelchairs, a form of powered wheelchair that integrates advanced software and hardware to safely navigate complex environments. Such technological advances not only enhance user capabilities in everyday tasks but also promote independence and accessibility. Despite being a research area that has been investigated for decades, multiple open questions and potential enhancements remain. We differentiate between traditional non-holonomic wheelchairs, with 2 degrees of freedom (DOFs), and holonomic (omnidirectional) wheelchairs with 3 DOFs. 

\begin{figure}
    \centering
    \includegraphics[width=\linewidth]{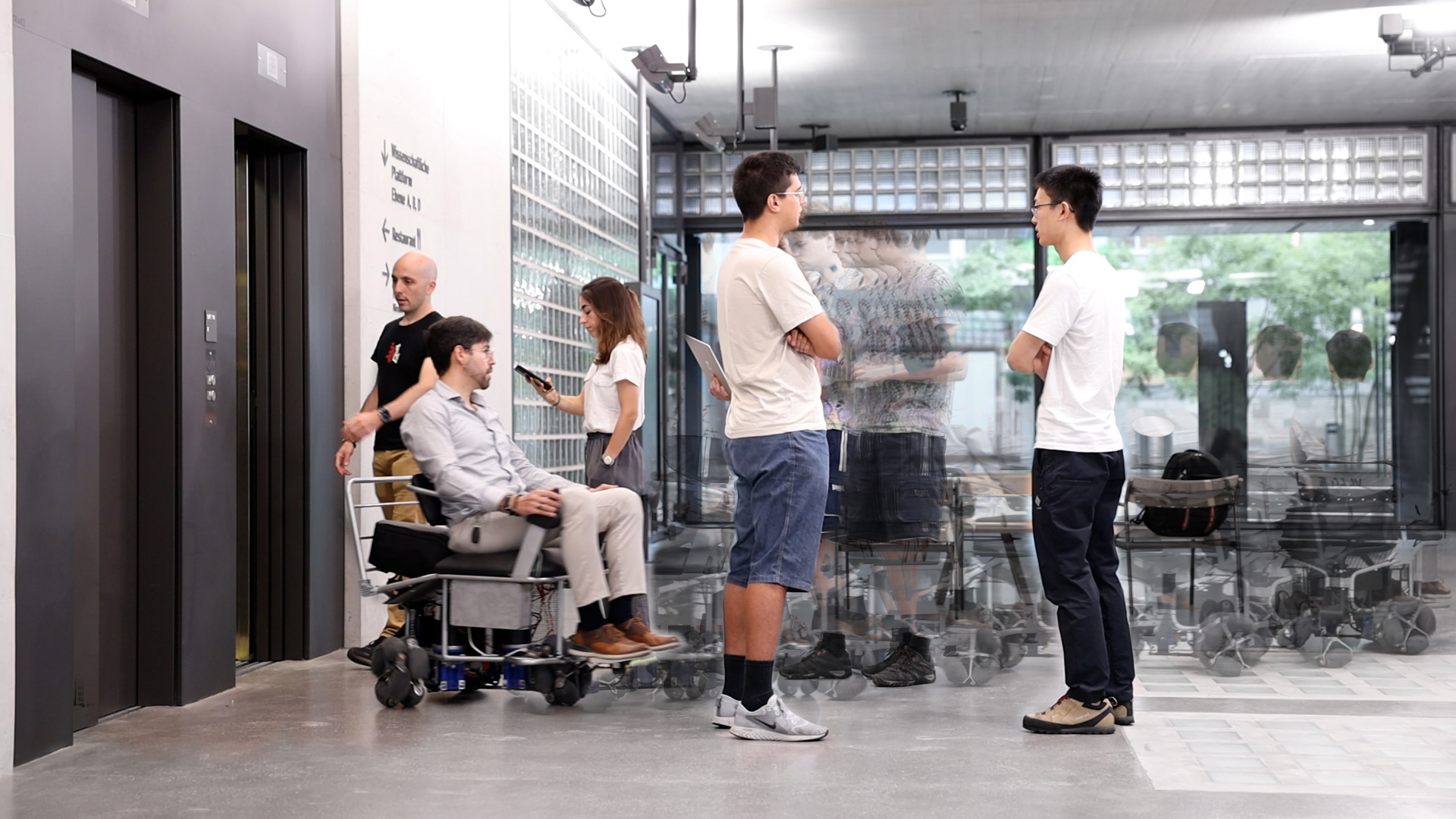}
    \caption{Illustration of the holonomic wheelchair used in this study.}
    \label{fig:daav_ethz}
\end{figure}

We focus on smart wheelchair navigation, which can be achieved in two main ways: fully autonomous systems take over full control of the wheelchair \cite{autonomous_wheelchair}, while shared control combines a user input with a planned motion \cite{shared_control_wheelchair_ex1, shared_control_wheelchair_ex2}. This is sometimes referred to as a semi-autonomous system. Not only can shared control enhance safety in navigation (e.g. for people with motion impairments), but also reduce cognitive load on the user, which leads to an improved user experience. Shared control can support the user in navigation and obstacle avoidance and reduce the controller space of the user, by taking over full control over certain degrees of freedom \cite{shared_control_robotic_arms}. Furthermore, it is preferable to fully autonomous systems, because it allows for more user authority. 

Although omnidirectional driving has much potential to enhance user experience, most current work focuses on safe (collision-free) navigation for non-holonomic (2 DOFs) wheelchairs \cite{rds_convex_nh}. Holonomic (3 DOFs) wheelchairs have so far received less attention from the research community and current approaches often lead to uncomfortable or unintuitive behavior from user perspective.

In this paper, we aim to develop a novel shared control method based on reinforcement learning (RL) for a holonomic (omnidirectional) wheelchair. We combine a user input provided by a joystick and 360° LiDAR data to ensure safe collision-free navigation. We further reduce the user input to 2D to reduce cognitive load on the user, while the wheelchair can move in all three degrees of freedom. Besides safe navigation, we also focus on user comfort and intuitive behavior from user perspective.

The main contribution of this work is the exploration of the applicability of reinforcement learning for shared control, e.g. by comparing different RL agents trained with varying reward functions. Additionally, we investigate the influence of different network architectures by adding a Convolutional Neural Network (CNN) to process the LiDAR readings and improve the agent's capability of differentiating between different obstacles \cite{cnn_for_maplesss_navigation}. We also include long short-term memory (LSTM) \cite{lstm} to improve the planning capability of the agent~\cite{lstm_for_memory}. We further introduce a new metric for cognitive load, which we call \emph{heading ($\Phi$)}. It is defined as the angle between robot orientation and the direction in which the driver is trying to go. By optimizing the robot movement to minimize $\Phi$, we minimize the amount of joystick corrections the user must perform while driving leading to a more comfortable and intuitive behavior from user perspective. At last, we also perform real-life experiments on a holonomic mobility platform of DAAV\footnote{\url{https://www.daav.ch/}}.

The rest of this paper is structured as follows: At first, we explore related works in reinforcement learning in robotics and non-learning and learning-based methods for shared control (\Cref{sec:related_works}). Then we provide more background information about the simulator and the RL algorithm (\Cref{sec:background}) used for training. Afterward, we dive into the actual shared control task we implement during training, in which we elaborate on the goal of the task, how we simulate a user input and describe the different training environments (\Cref{subsec:shared_control_task}). Furthermore, we also describe observation and action space (\ref{subsec:obs_ac}), the reward functions (\Cref{subsec:reward}), and the different model architectures used for the agent (\Cref{subsec:models}). The training setup (hardware and hyper-parameters). together with the experimental setup and results are discussed in detail in \Cref{sec:results} followed by the conclusions and an outlook on future work (\Cref{sec:conclusion}).

%% file: sec/02_RelatedWorks.tex

\section{Related Works}
\label{sec:related_works}
\subsection{Reinforcement Learning for Autonomous Systems}
Most work regarding RL and robotics controls focuses on autonomous systems. Some examples include legged locomotion~\cite{rl_legged_locomotion}, wheeled-legged locomotion~\cite{rl_wheeled_legged}, humanoids~\cite{rl_for_humanoids}, drone racing~\cite{rl_drone_racing} or even large machines such as excavators~\cite{rl_for_excavator}. Other work also focuses on using RL for autonomous navigation (\cite{rl_crowds_aut1, rl_crowds_aut2, rl_crowds_aut3}. For those systems, the task is usually to reach a goal in space. Therefore, the goal position or orientation can be included in the observation as well as the reward function. This is not optimal for shared control, in which we do not want to reach a goal, but track the user input as exactly as possible while ensuring safe navigation. A method to solve a similar issue is presented in~\cite{walktheseways}, in which the goal is to follow velocities in legged robot locomotion, however, without avoiding obstacles.

In wheelchair navigation, we use 2D rotations. In~\cite{orient_rep}, it is shown that a 2D representation $\alpha$ should be represented as $[cos(\alpha), sin(\alpha)]^T$ to avoid discontinuities in the function $g:SO(2)\rightarrow \alpha$, which can influence the learning process.

\subsection{Shared Control}
\textbf{Non-Learning Based Methods.}
Most existing solutions for shared control are not learning-based and so-called reactive methods.~\citeauthor{rds_convex_nh} propose Reactive Driving Support (RDS), a method which constructs Velocity Obstacles~\cite{vo} between obstacles and the robot (\cite{rds_convex_nh, rds_evaluation_metrics}). The robot velocity is then calculated to be in the safe velocity space while minimizing the cost $|\textbf{v}_{user}-\textbf{v}_{robot}|^2$. RDS is originally designed for non-holonomic robots ($\textbf{v}=(v_x, \omega)^T$). In our laboratory, an initial implementation also exists for holonomic systems (($\textbf{v}=(v_x, v_y, \omega)^T$)), for which weights were set to balance lateral and angular velocity in the optimization function. However, we observed that this does not guarantee optimal behavior in all situations and can lead to unintuitive behavior from user perspective.

\textbf{Learning Based Methods.}
Compared to autonomous systems, research regarding shared control is rather limited.~\cite{shared_autonomy_via_drl} proposes a method of shared autonomy by decomposing the reward function into two parts: The first part focuses on navigation requirements, such as reaching a goal or obstacle avoidance, while the second part captures user feedback. 

However, the proposed method requires discrete human input during training, which could be unsuitable for continuous input like in wheelchair navigation. To solve that issue,~\citeauthor{rlbussc} propose a method based on~\cite{shared_autonomy_via_res_policy_rl} and use behavior cloning combined with Generative Adversarial Imitation Learning~\cite{GAIL} to train a user policy to provide continuous input during training~\cite{rlbussc}. It is used to train a non-holonomic wheelchair to navigate through crowds. However, during training the wheelchair must reach a goal, whose position is used in the state provided to the agent and in the reward function. This could lead to problems in real-life applications because the goal of the user is unknown. Furthermore, the approach uses agent-level data of the surroundings, rather than raw sensor data, which prohibits tests with a real system.

%% file: sec/03_Background.tex

\section{Background}
\label{sec:background}

\subsection{Isaac Gym}
Isaac Gym was specifically developed for RL and is built on top of the Isaac Simulator, which is powered by NVIDIA~\cite{isaac_gym}. We chose the simulator based on its high parallelization and extensive GPU use for both policy training and task optimization. The simulator also already supports different robot models (non-holonomic \& holonomic) and provides implementations of low-level controllers, which allows us to apply the output of our agent directly. During training, we use NVIDIA Kaya \footnote{\url{https://github.com/nvidia-isaac/kaya-robot?tab=MIT-1-ov-file}}, a holonomic robot, which has the same three-wheel structure as our wheelchair.

\subsection{Reinforcement Learning \& PPO}
Generally, robot navigation problems can be formulated as Markov Decision Processes (MDP,~\cite{rl_mdp}), which consist of a set of states $S$, actions $A$, transmissions $T$, and a reward function $R$.  A requirement of an MDP is, that the state $S$ is fully observable. This is not the case for shared control, in which the goal or intention of the user is not measurable. In this case, the problem can be formulated as a Partially Observed Markov Decision Process (POMDP,~\cite{pomdp}), in which part of the state is not observable but can be estimated via RL.

In this work, we are using proximal policy optimization (PPO,~\cite{ppo}), which is a popular RL algorithm and has been shown to be successful for robot navigation tasks~\cite{ppo_robot_navigation}. PPO ensures stable and reliable training due to its clipped objective function, which prevents large policy updates during training. Furthermore, it is sample efficient, parallelizable, and robust to the choice of hyperparameters~\cite{ppo, ppo_hparams}. It allows for a continuous action space and high dimensional observations, which is crucial for our task since we use LiDAR data as an input~\cite{ppo_continuous_control}.

%% file: sec/04_Methodology.tex

\section{RL for Shared Control}
\label{sec:methodology}
\subsection{Shared Control Task}
\label{subsec:shared_control_task}
\begin{figure}[t]
\begin {center}
\includegraphics[width=0.5\textwidth]{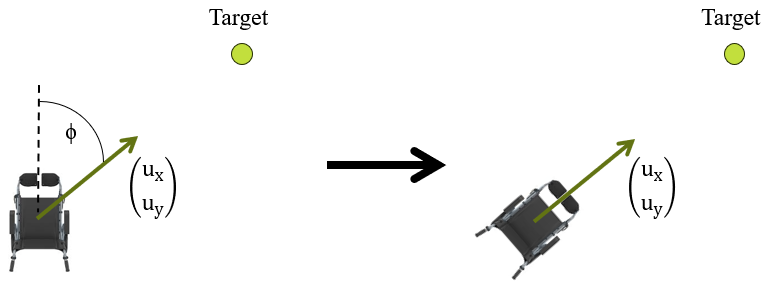}
\caption{The user input always points towards the target, while the heading ($\Phi$) is minimized at all times.}
\label{fig:heading}
\end {center}
\end{figure}

\textbf{Goal}. The main goal of the training task is to reach a target point in 2D space. The target is only fictional, meaning that it is only used to simulate the user input and no information about it is provided to the agent. Furthermore, we made a design decision, that the wheelchair should always face the target as well as possible. This is called \emph{heading ($\Phi$)} and describes the angle between the yaw of the wheelchair and the angle towards the target (\Cref{fig:heading}). 

\textbf{User Model.} As mentioned above, during training there is a fictional target. It is assumed that the user is always pointing the joystick $(u_x, u_y)^T$ towards said target at any time. During training, the vector for the user input is normalized to have norm one. Note that in test time the user input can have different norms depending on the extension level of the joystick.

\begin{figure}[t]
\begin {center}
\includegraphics[width=0.42\textwidth]{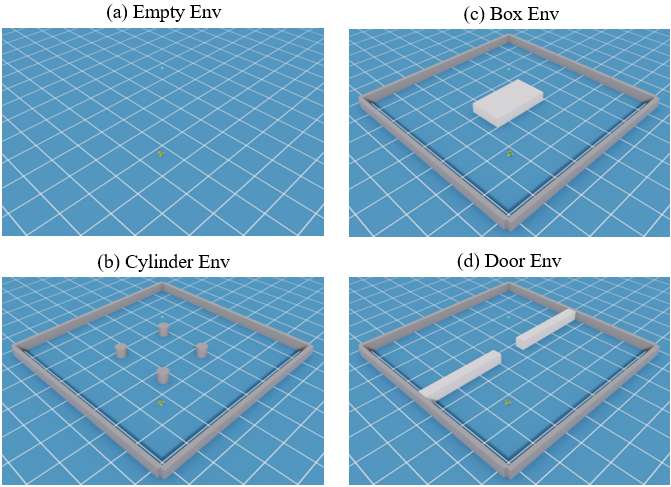}
\caption{We use four different training environments in Isaac Gym (dark obstacles are of static size, light obstacles are of varying size).}
\label{fig:envs}
\end {center}
\end{figure}

\textbf{Training Environments.} During training we use four different environments (\Cref{fig:envs}):

\begin{enumerate}[a)]
\item \emph{Empty Environment.} The empty environment does not contain any obstacles. At reset, the starting position and orientation of the robot as well as the target position are randomized.

\item \emph{Cylinder Environment.} The second environment contains four cylinders that are positioned asymmetrically. The cylinders are of static size, even after a reset. The randomization at reset works the same as for the empty environment.

\item \emph{Box Environment.} The box environment has one large obstacle in the middle of the space. The box is of random size ($l\in[1.0m, 4.0m]$, $b\in[1.0m, 2.0m]$) and the size is randomly set at reset. The main goal of this environment is for the agent to plan and learn to move around large obstacles of varying sizes. The starting position and target position always lie on the opposite side of the obstacle.

\item \emph{Door Environment.} The door environment contains only a door in the middle of the space, which is of varying width: $w\in[0.9m, 1.75m]$. The starting position is randomized on one side of the door, while the target is set to be in the opposing quadrant of the environment, such that it is on the other side of the door while the initial user input should point somewhat towards the door no matter the starting position. This task aims to train the agent to pass tight spaces and it is the only asymmetric environment out of the four.
\end{enumerate}

\subsection{Observation \& Action}
\label{subsec:obs_ac}
The model's input represents the agent's observation, i.e the concatenation of agent state, user input and sensor data:

\vspace{6pt}
\begin{tabular}{ll}
Lidar Ranges:   & $(d_1, ..., d_n)^T$ \\
User Input:     & $(u_{x}, u_y)^T$ \\
Base Velocity:  & $(v_{x\_meas}, v_{y\_meas}, \omega_{meas})^T$ \\
Last Action: & $(v_{x,t-1}, v_{y, t-1}, \omega_{t-1})^T$ \\
$2^{nd}$ Last Action: & $(v_{x,t-2}, v_{y, t-2}, \omega_{t-2})^T$
\vspace{6pt}
\end{tabular}

The LiDAR ranges consist of $n$ ranges equally distributed $360$° around the robot. We use either $36$ or $360$ samples depending on the model architecture. They are normalized with the maximum LiDAR range, which is constant $360$° around the robot. The user input is a vector in the direction in which the user wants to go ($u_x, u_y \in [0, 1]$). Furthermore, the current velocity measurements of the robot as well as the last two actions are provided as well. All velocity inputs are normalized corresponding to the maximum velocity of the robot. 

The \emph{Action} represents the output of the model and corresponds to the new velocity command given to the robot:
\begin{equation}
Action = (v_{x}, v_{y}, \omega)^T  
\end{equation}

The action space is $[0, 1]$, which is why the network output is scaled with the maximum linear and angular velocity respectively before being passed to the robot.

\subsection{Reward \& Collision Model}
\label{subsec:reward}

We use two different reward functions consisting of different parts which are summarized in \Cref{tab:reward}.

\begin{table}[]
\begin{center}
\caption{Reward Function: \tikzcircle[fill=green]{3pt}: Method 1\&2, \tikzcircle[fill=blue]{3pt}: Method 1, \tikzcircle[fill=red]{3pt}: Method 2, $r_*$: reward weight, $d_*$: collision model parameter, $d$: LiDAR range}
\label{tab:reward}
\begin{tabular}{ll}
\hline
\textbf{Term}                      & \textbf{Equation} \\ \hline
\tikzcircle[fill=green]{3pt} Obstacles       &   $\left\{\begin{matrix}
                              r_c + r_{crit} * (d_{crit} - d)^2,   & d < d_{crit} \\
                              r_{col},   & d < d_{col}  \\
                                0 &  else\\
                                \end{matrix}\right. $ \\
\tikzcircle[fill=green]{3pt} Heading         & $\left\{\begin{matrix}
                              r_h * |\Phi|^2,   & |\Phi| > \Phi_{treash} \\
                              0,   & else  \\
                                \end{matrix}\right. $\\
\tikzcircle[fill=blue]{3pt} $v_{xy}$ Tracking  &  $r_a * exp(r_l * |v_{xy} - u_{xy}|^2)$\\
\tikzcircle[fill=red]{3pt} $v_{x}$ Tracking  &  $r_a * exp(r_l * |v_{x} - u_{x}|^2)$\\
\tikzcircle[fill=red]{3pt} $v_{y}$ Punishment & $r_{v_y} * |v_y|^2$ \\
\tikzcircle[fill=green]{3pt} Smoothing         &  $r_{as} * |a_t - a_{t-1}|^2$\\
\tikzcircle[fill=green]{3pt} Smoothing $2^{nd}$ order &  $r_{as} * |a_t - 2*a_{t-1} + a_{t-2}|^2$ \\ \hline \\
\end{tabular}
\end{center}
\end{table}

As visible, there are some parameters related to the collision model. As mentioned in \Cref{subsec:obs_ac}, the LiDAR data is a vector of $n$ ranges. We assign two collision model parameters for each LiDAR range: $d_{col}$ and $d_{crit}$ with $d_{crit} > d_{col}$. If the LiDAR range ($d$) is smaller than $d_{crit}$ there is a collision and the task and the environment are reset during training. If $d_{crit} > d > d_{col}$ we impose a punishment in the reward function to keep the robot from getting too close to obstacles. The chosen collision model has the shape of a capsule with the center being at the center of the robot. Every LiDAR range is assigned its own $d_{col}$ \& $d_{crit}$ depending on its position around the robot. The RL approach also allows other collision model shapes such as circles or squares.

The difference between the two methods lies in the user intention tracking. For method 1 we maximize the match of user input $(u_x, u_y)^T$ to the linear base velocity of the robot $(v_x, v_y)^T$. For method 2, we only match $u_x$ to $v_x$ and punish any movements in lateral direction. This ensures that the robot only moves sideways if necessary, e.g. to avoid a collision. We further punish the \emph{heading}, however, only if it is larger than a threshold to avoid unstable behavior for small values. At last, we include action smoothing terms to ensure the smoothness of consecutive actions.


\subsection{Network Architectures}
\label{subsec:models}

\begin{figure}[]
\begin {center}
\includegraphics[width=0.48\textwidth]{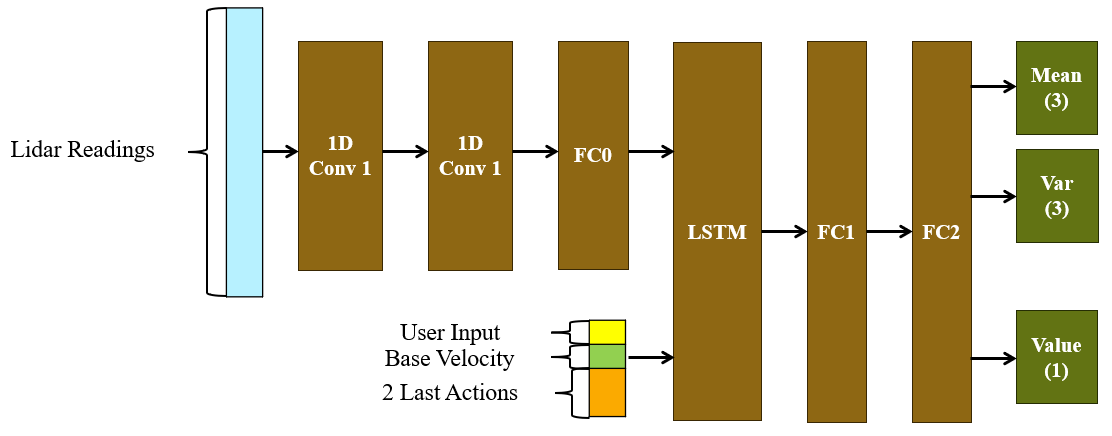}
\caption{General network architecture with layer names.}
\label{fig:architecture}
\end {center}
\end{figure}

We use different network architectures, which all have a similar structure. The main parts are a LiDAR CNN (LCNN) and an RNN (LSTM) followed by fully connected layers (FCN). The LCNN processes the LiDAR data for feature extraction and the LSTM is used to increase planning capabilities. The general structure is shown in Fig. \ref{fig:architecture}, however, the models differ in the number of units and the actual parts that are implemented. For networks without LCNN, the LiDAR ranges are concatenated with the rest of the observation and directly passed into the first layer.

The models that we trained differ in the network architecture, the reward function (method and/or weights) and the environments used during training. The reward weights are summarized in \Cref{tab:rew_params} and the network architectures and training environments in \Cref{tab:architectures}.

\begin{table*}[t]
\begin{center}
\caption{Reward Weights: If there is no value for $r_{v_y}$ method 1 is used, otherwise method 2.}
\label{tab:rew_params}
\renewcommand{\arraystretch}{1.4}
\begin{tabular}{cccccccccc}
\toprule
\multirow{3}{*}{\textbf{\begin{tabular}[c]{@{}c@{}}Model \end{tabular}}} & \multicolumn{9}{c}{\textbf{Reward Weights}}                                                                                   \\ 
& \multicolumn{3}{c}{\textbf{Obstacles}}                                 & \multicolumn{2}{c}{\textbf{Heading}}            & \multicolumn{3}{c}{\textbf{Velocity Tracking}}                        & \textbf{Smoothing}    \\ \cline{2-10} 
& $r_c$                   & $r_{crit}$                & \multicolumn{1}{c}{$r_{col}$} & $r_h$                   & \multicolumn{1}{c}{$\Phi_{thresh}$} & $r_a$                   & $r_l$                   & \multicolumn{1}{c}{$r_{v_y}$} & $r_{as}$                   \\ 
\midrule
FC, LFC                                                                                  & -2                    & -10                    & \multicolumn{1}{c}{-100}     & -0.2                     & \multicolumn{1}{c}{0.2}     & 0.5                     & -0.5                     & \multicolumn{1}{c}{-}    & -0.02                      \\
CLFC, CLFC\_D, SCLFC\_D\_R1                                                                                 & -1                     & -1                    & \multicolumn{1}{c}{-100}     & -0.2                     & \multicolumn{1}{c}{0.2}     & 0.5                     & -0.5                     & \multicolumn{1}{c}{-}    & -0.02                      \\
SCLFC\_D\_R2                                                                                  & -1                     & -1                     & \multicolumn{1}{c}{-100}     & -0.5                     & \multicolumn{1}{c}{0.2}     & 0.5                     & -0.5                     & \multicolumn{1}{c}{-1.6}    & -0.02                      \\
\bottomrule
\end{tabular}
\end{center}
\end{table*}

\begin{table*}[]
\begin{center}
\caption{Detailed model architectures and environments used during training: For 1D convolutional layers, the numbers represent (kernel size, channels, stride).}
\label{tab:architectures}
\renewcommand{\arraystretch}{1.4}
\begin{tabular}{ccccccccccc}
\toprule
\multirow{3}{*}{\textbf{\begin{tabular}[c]{@{}c@{}}Model\end{tabular}}} & \multicolumn{6}{c}{\textbf{Architecture}}      & \multicolumn{4}{c}{\textbf{Evironments}}                                                  \\ 
& \multicolumn{3}{c}{\textbf{LCNN}}               & \textbf{LSTM} & \multicolumn{2}{c}{\textbf{FC}} & \multicolumn{1}{c}{\multirow{2}{*}{\textbf{a}}} & \multicolumn{1}{c}{\multirow{2}{*}{\textbf{b}}} & \multicolumn{1}{c}{\multirow{2}{*}{\textbf{c}}} & \multirow{2}{*}{\textbf{d}} \\ \cline{2-7} 
& 1D Conv 1 & 1D Conv 2 & \multicolumn{1}{c}{FC0} & LSTM                              & FC1             & FC2            & \multicolumn{1}{c}{}                            & \multicolumn{1}{c}{}                            & \multicolumn{1}{c}{}                            &                             \\ 
\midrule
FC                                                                                   & -          &  -         & \multicolumn{1}{c}{-}    & \multicolumn{1}{c}{-}                               & 64                & 64               & \CheckmarkBold                                                 &  \CheckmarkBold                                                & \CheckmarkBold                                                 &  \XSolidBrush                           \\
LFC                                                                                & -          & -          & \multicolumn{1}{c}{-}    & \multicolumn{1}{c}{128}                               & 64                & 64               &  \CheckmarkBold                                                &  \CheckmarkBold                                                & \CheckmarkBold                                                 & \XSolidBrush                             \\
CLFC                                                                                   & (5,16,2)          & (3,8,2)          & \multicolumn{1}{c}{128}    & \multicolumn{1}{c}{128}                               & 128                & -               &  \CheckmarkBold                                                &   \CheckmarkBold                                               &  \CheckmarkBold                                                &  \XSolidBrush                           \\
CLFC\_D                                                                        & (5,16,2)          & (3,8,2)          & \multicolumn{1}{c}{128}    & \multicolumn{1}{c}{128}                               & 128                & -               &  \CheckmarkBold                                                &   \CheckmarkBold                                               &  \CheckmarkBold                                                &  \CheckmarkBold                           \\
SCLFC\_D\_R1/R2                                                                         & (5,8,2)          & (3,4,2)          & \multicolumn{1}{c}{64}    & \multicolumn{1}{c}{64}                               & 64                & -               &  \CheckmarkBold                                                &   \CheckmarkBold                                               &  \CheckmarkBold                                                &  \CheckmarkBold                           \\ 
\bottomrule
\end{tabular}
\end{center}
\end{table*}

%% file: sec/05_Experiments.tex

\section{Experimental Results}
\label{sec:results}


\subsection{Experimental Setup: Simulation \& Training}
\label{sec:simulation_training}

\subsubsection{Simulation Specifications}
The simulation is implemented in Isaac Sim version 2022.2.1. The machine used for simulation and training is equipped with 13$^{th}$ Gen Intel{\textregistered} Core{\texttrademark} i9-13900K and graphics card NVIDIA GeForce RTX 4090.

For the training process, we adopt a curriculum learning approach. At first, we only train the agent to reach a goal in empty space without obstacles (empty environment) to train how to track the user input. We found that if we later on only include environments with obstacles, the agent loses simple skills such as just driving straight. Therefore, in the second step, we include all environments (empty plus a selection of the ones with obstacles). The different environments used during training are distributed evenly over the total amount of environments. We train the first part of the curriculum for 50 epochs, while the total training of the policy converges after approximately 300 epochs.

\textbf{Isaac Gym Parameters.} We run training with a total of 128 environments. The control frequency is the same as the frequency with which the simulator is updated, which is 40Hz.

\textbf{PPO Hyperparameters.} 
PPO is implemented with a learning rate of 5e-4, the entropy coefficient is set to 1e-2 and $\epsilon_{clip}$ is set to 0.2. We train with a horizon length of 128, a minibatch size of 4096 and 4 mini epochs.

\subsubsection{Wheelchair Specifications}
The maximum linear and angular velocities during training are set to $1\frac{m}{s}$ and $1\frac{rad}{s}$ respectively. However, we noticed some discrepancies in the commanded velocity and measured velocity for Kaya in Isaac Sim, which is why during evaluation the maximum linear and angular velocity are adjusted respectively (\Cref{sec:results}). We further set the LiDARs to have a minimum range of $0.3m$ and a maximum range of $2.5m$.

The collision model is designed as a capsule with two half circles with a diameter of $0.65m$ for $d_{col}$ and $1.05m$ for $d_{crit}$ spaced by $0.3m$.


\subsection{Simulation Results (Gazebo)}

We use the Ridgeback robot\footnote{\url{https://github.com/ridgeback/ridgeback_simulator}} for simulations in Gazebo. Due to the velocity inaccuracies of Kaya in Isaac Sim, the maximum linear velocity was reduced to $0.67\frac{m}{s}$ and the maximum angular velocity was set to $2\frac{rad}{s}$.

\begin{figure*}[t]
    \minipage{0.495\textwidth}
      \includegraphics[width=\linewidth]{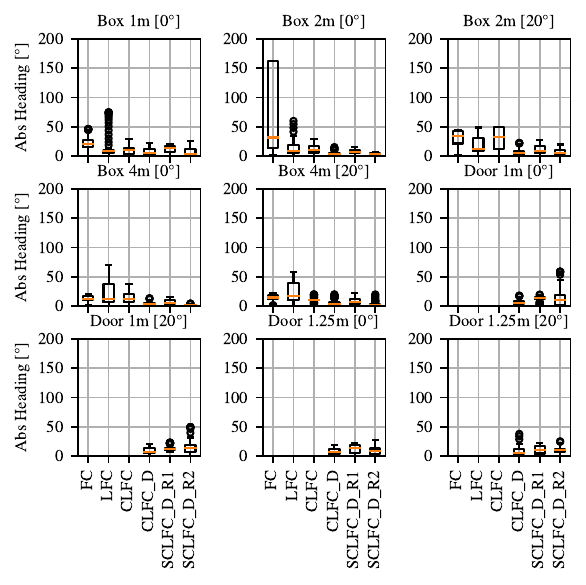}
    \endminipage\hfill
    \minipage{0.495\textwidth}%
      \includegraphics[width=\linewidth]{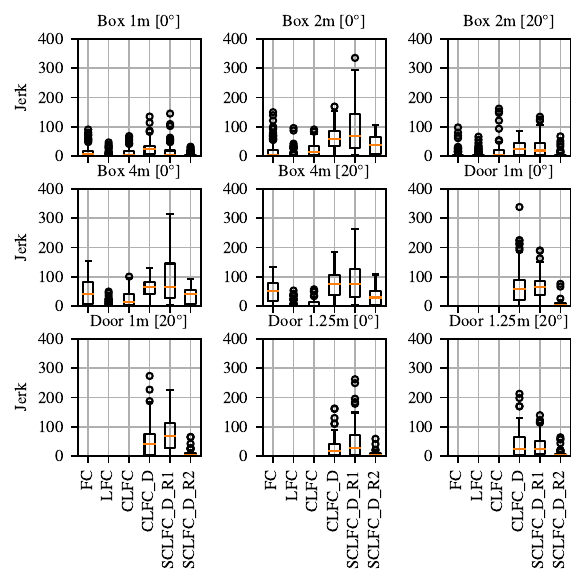}
    \endminipage
    \caption{We evaluate our agents based on heading and jerk (as defined in the text). The titles of the subplots refer to the task name, obstacle size and incident angle.}
    \label{fig:model_comp}
\end{figure*}

\begin{figure*}[t]
    \minipage{0.32\textwidth}
      \includegraphics[width=\linewidth]{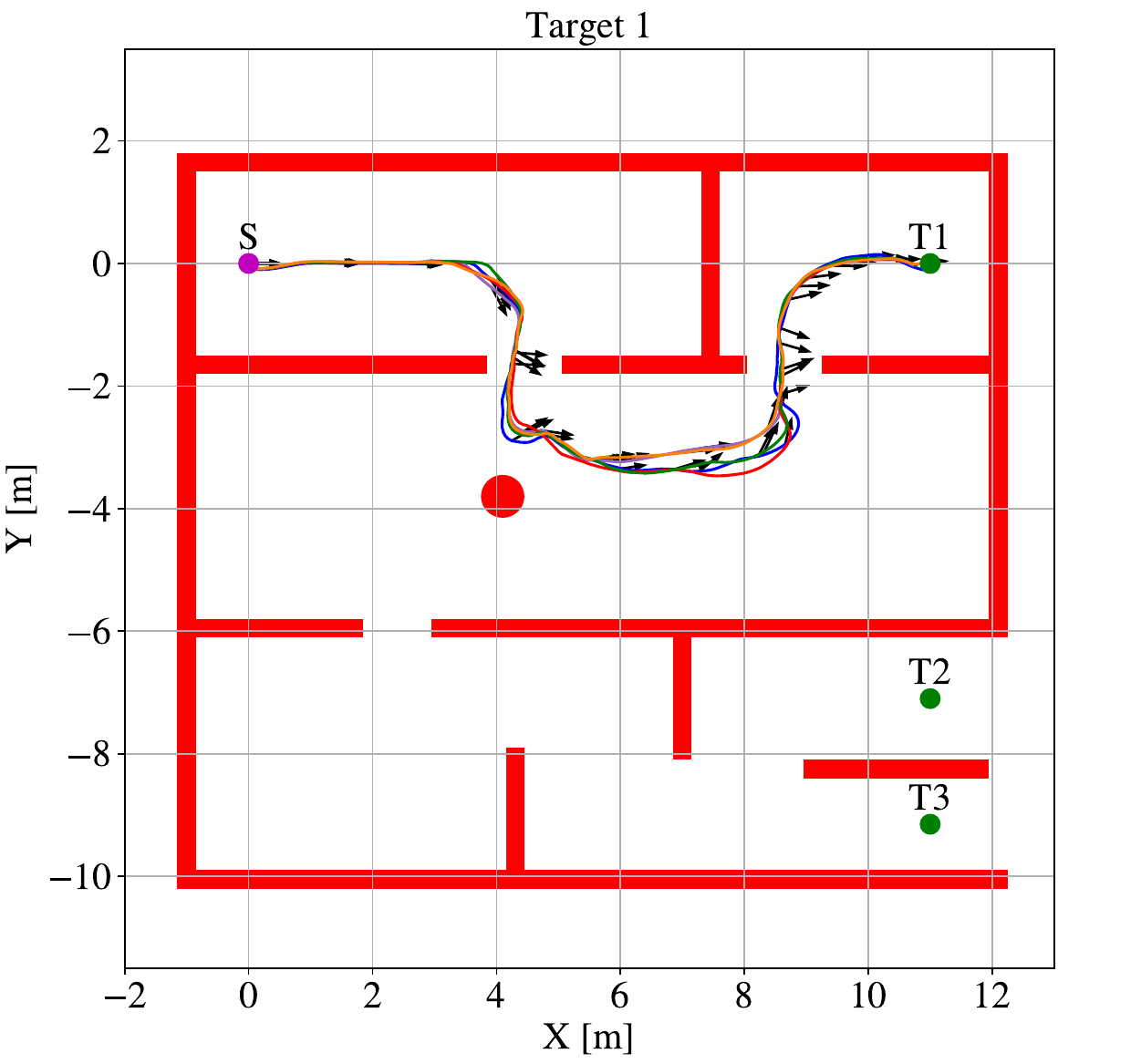}
    \endminipage\hfill
    \minipage{0.32\textwidth}
      \includegraphics[width=\linewidth]{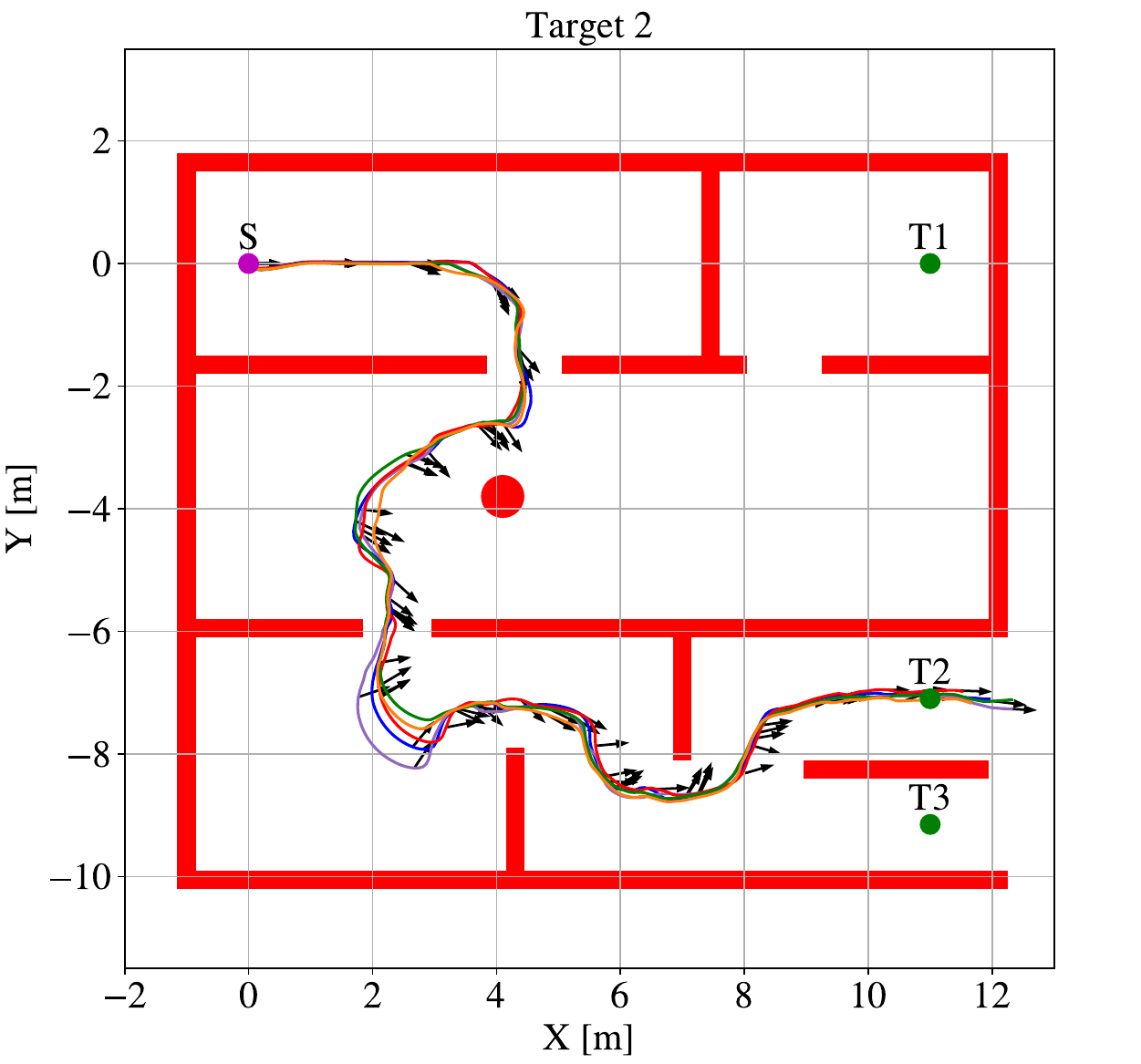}
    \endminipage\hfill
    \minipage{0.32\textwidth}%
      \includegraphics[width=\linewidth]{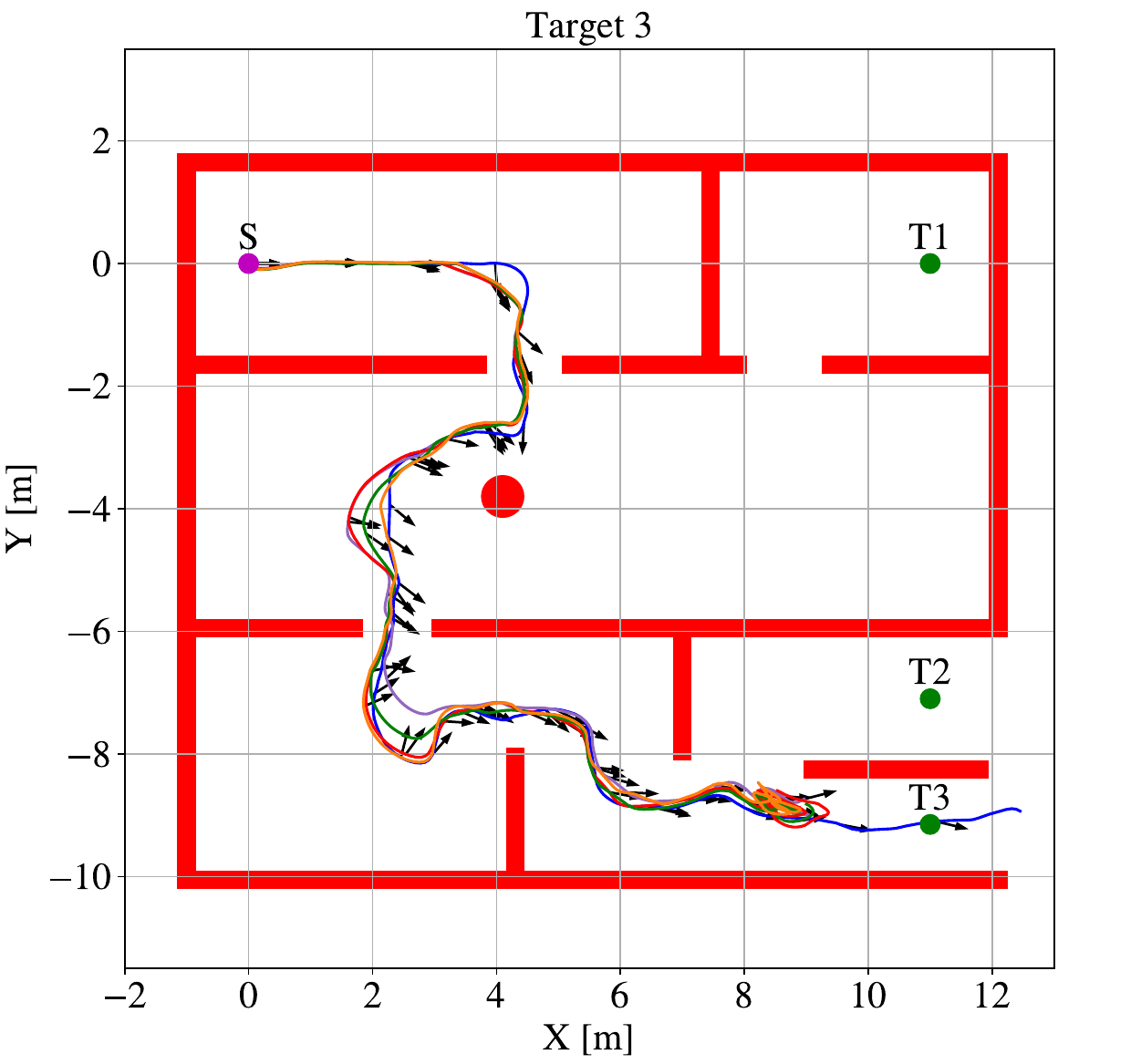}
    \endminipage
    \caption{We conduct a qualitative analysis in an unseen environment with human user input. The goal is to reach three different targets 5 times. The plots show trajectories and the user input at given points (black arrows).}
    \label{fig:manual}
\end{figure*}

\subsubsection{Simulated User Input}
\label{subsec:simulated_user_input}
To compare the different models presented in \Cref{tab:rew_params} and \Cref{tab:architectures}, experiments similar to the training tasks are conducted in two main setups: First the performance is measured by navigating around a growing obstacle (box) $1m$ wide with lengths of $1m$, $2m$ and $4m$. Second, the performance is measured by passing doors of two sizes $1m$ and $1.25m$. The target position is set directly behind the obstacle ($2m$), while the starting position is set $2m$ in front of the obstacle with an incident angle of either $0$° or $20$° in relation to the target. The user input is simulated as during training and is always pointing towards the target.

The models are compared on two main metrics: The heading (facing towards the target) and the jerk of the robot velocity $(v_x, v_y, \omega)^T$ (smoothness). The heading is calculated based on the user input as $\Phi = |\arctantwo(\frac{u_{y}}{u_{x}})|$ and is used as a metric for cognitive load. Jerk is used to determine trajectory smoothness and evaluate user comfort ($J=\sqrt{J_{x}^2+J_{y}^2+J_\theta^2}$). The results are summarized in Fig. \ref{fig:model_comp}. The absence of a plot for a specific model indicates that it was not able to successfully solve the task and that the robot got stuck, e.g. in unnecessary movement to avoid collision.

Generally, models are only able to solve tasks they were trained in. None of the models trained in all except the door tasks can pass a door. Regarding planning capabilities (navigating around a large obstacle), it is visible that the usage of LSTM generally improves performance. However, including all tasks in the training process reduces planning capabilities compared to the case when the door task is left out. Also, the usage of an LCNN seems to slightly reduce planning capabilities. An attempt was made to freeze the LCNN at some point during training to keep the input for the LSTM constant, however, we did not observe any improvement. For the two model sizes used, the improvement observed for the larger model is rather small and mostly regarding the heading performance and not trajectory smoothness. The two reward methods show a major difference regarding trajectory smoothness, with the jerk of method 2 being much smaller. Furthermore, SCLDC\_D\_R2 shows the best performance in the door task, being able to pass a space with a $1m$ width. Going through tight spaces is prioritized over moving around large obstacles as it is more commonly faced in real life. Therefore, SCLDC\_D\_R2 is chosen as the best-performing model and evaluated further in the following.

\begin{figure}[]
    \centering
    \minipage{0.35\textwidth}
        \centering
      \includegraphics[width=\linewidth]{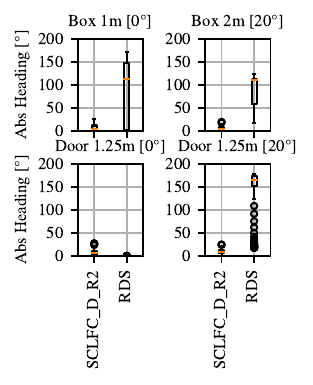}
    \endminipage
    
    \minipage{0.35\textwidth}
        \centering
      \includegraphics[width=\linewidth]{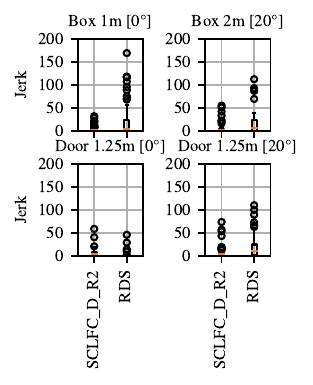}
    \endminipage
    \caption{To compare the RL approach to RDS, we use the same setup as in \ref{fig:model_comp}.}
    \label{fig:rds_comp}
\end{figure}

Figure \ref{fig:rds_comp} shows a comparison of the model to RDS \cite{rds_convex_nh} for a selection of the experiments mentioned before. Generally, the heading performance of the RL approach is superior. However, note that the user input is 2D, which is not what RDS is designed for as it has a 3D input. Therefore, once RDS turns away from the target, it never turns back to face it in these experiments. Regarding trajectory smoothness, the RL approach shows competitive or even superior performance with smaller jerk and fewer fliers compared to RDS.

\subsubsection{Manual User Input}
\label{subsec:manual_user_input}

This experiment aims to qualitatively evaluate the model in an environment it has not seen during training. Furthermore, the user input is provided manually with a physical joystick and a camera is added on top of the robot in the simulation to allow for point-of-view driving. The environment as well as the starting point and targets of the task are depicted in Fig. \ref{fig:manual}. Three different targets are defined: To reach T1, the driver must enter a room and navigate to the end. For T2 and T3, the driver must navigate through several rooms and finally pass a corridor with widths $2m$ and $1.5m$ respectively. Each task was tackled five times in total. For T1 and T2 each trial was successful, while for T3 only one out of five trials reached the final target because the robot started to retract after entering the corridor. This shows that the chosen user model in simulation is eligible and the agent still ensures safe navigation even with a human input. Generally, the agent also works for unseen environments and tasks. However, when the environment differs too much from training environments, e.g. a long tight corridor, the behavior can become unpredictable.

\subsection{Real World Validation}
\label{subsec:real_world_validation}
\begin{figure}[t]
    \minipage{0.23\textwidth}
      \includegraphics[width=\linewidth]{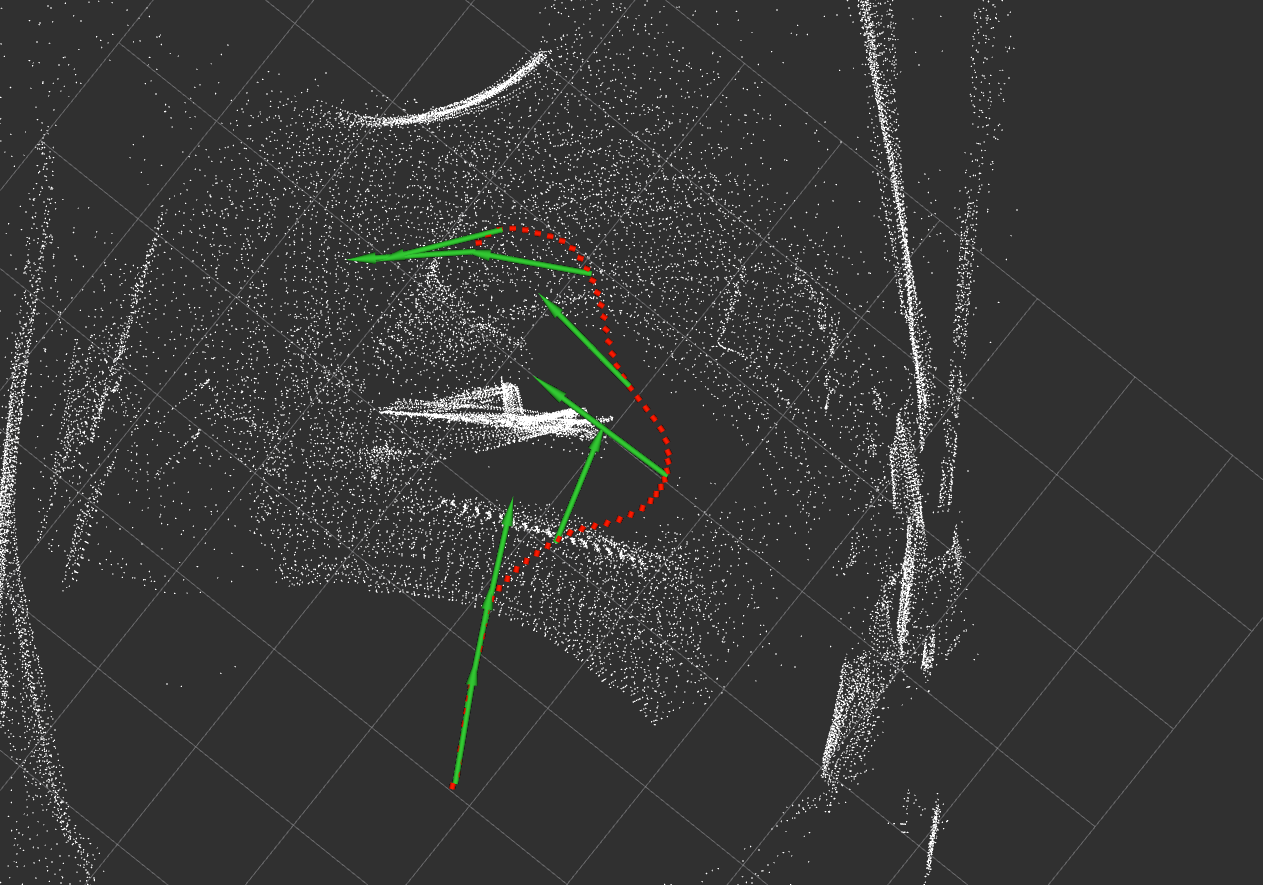}
    \endminipage\hfill
    \minipage{0.23\textwidth}
      \includegraphics[width=\linewidth]{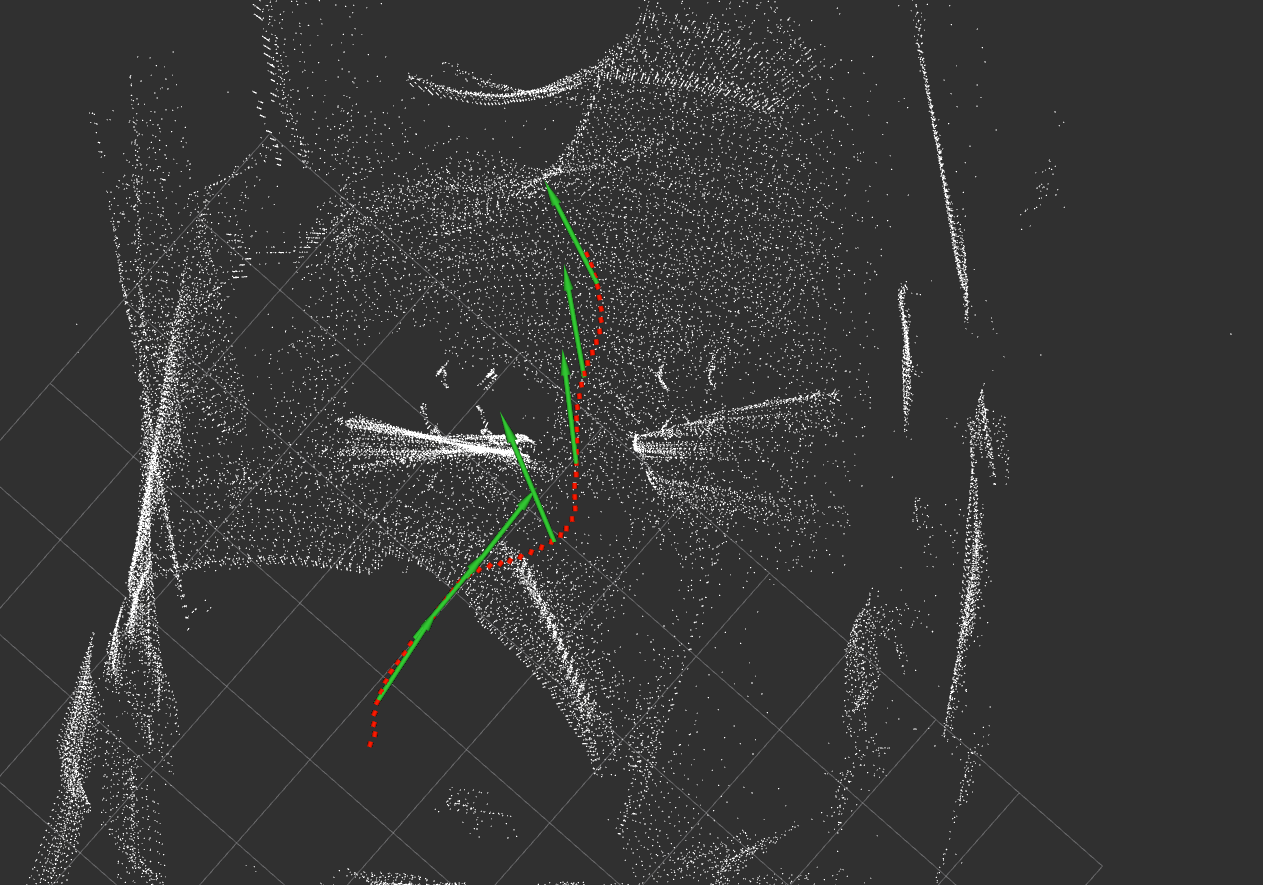}
    \endminipage
    \caption{Trajectories (red), user input (green), and LiDAR data (white) of real-world test runs. The figure on the left shows navigation around a box with $\sim1.65m$ width and the figure on the right shows navigation through a tight space ($\sim1m$) while approaching from the side.}
    \label{fig:real_world}
\end{figure}

Finally, to assess the transferability of results to a real robot, we performed a qualitative analysis on an omnidirectional wheelchair from DAAV. The setup was similar to the simulation with either a large box or a door-like tight space in front of the wheelchair. Figure \ref{fig:real_world} shows two examples of the tasks, showing the trajectories (red), user inputs (green arrows), and the LiDAR readings (white). The trajectory follows the user input, if there are no obstacles close to the wheelchair and deviates from it to avoid obstacles. The experiments show that the RL agent can be easily exported and integrated into the control loop of a real robot. Furthermore, the real wheelchair shows the same behavior as observed in simulation. 

%% file: sec/06_Conclusion.tex

\section{Conclusion}
\label{sec:conclusion}
In this paper, we successfully trained an agent using reinforcement learning that takes a 2D input and translates it to a 3D motion while optimizing user intention tracking, avoiding collisions and orienting the wheelchair in the proper direction. Our approach ensures collision-free navigation while optimizing user comfort through smoothness and reduced cognitive load goals. Regarding smoothness, we show competitive behavior to existing approaches. However, robustness regarding user intention tracking cannot be guaranteed for environments that differ too much from the training environments, which is a general problem of RL. Nevertheless, we show that the system can be easily transferred to a real omnidirectional wheelchair, which, to the best of our knowledge, has not been done in previous literature. We further demonstrate the potential of shared control for holonomic mobility platforms, which is covered in a limited amount of literature. At last, our approach allows various collision model shapes to cover both convex and non-convex shapes.

Based on the above, in future work, we would like to add more complexity and variety to our training environments to increase robustness. More complex environments also require a more complex user model, which imitates a human input. At last, we also want to include moving obstacles in the training process for crowd navigation.

The experiments on the real system were qualitative. Therefore, we will further evaluate the performance, by conducting more extensive studies including different participants.

%% file: sec/XX_Appendix.tex
\textcolor{white}{EOD}
\newpage
\onecolumn
\section*{Appendix: Further Results}

\vspace{2em}

\subsection{Trajectories of simulated user input experiments \ref{subsec:simulated_user_input}.}

\vspace{1.23em}

\noindent
    \minipage{0.32\textwidth}
      \includegraphics[width=\linewidth]{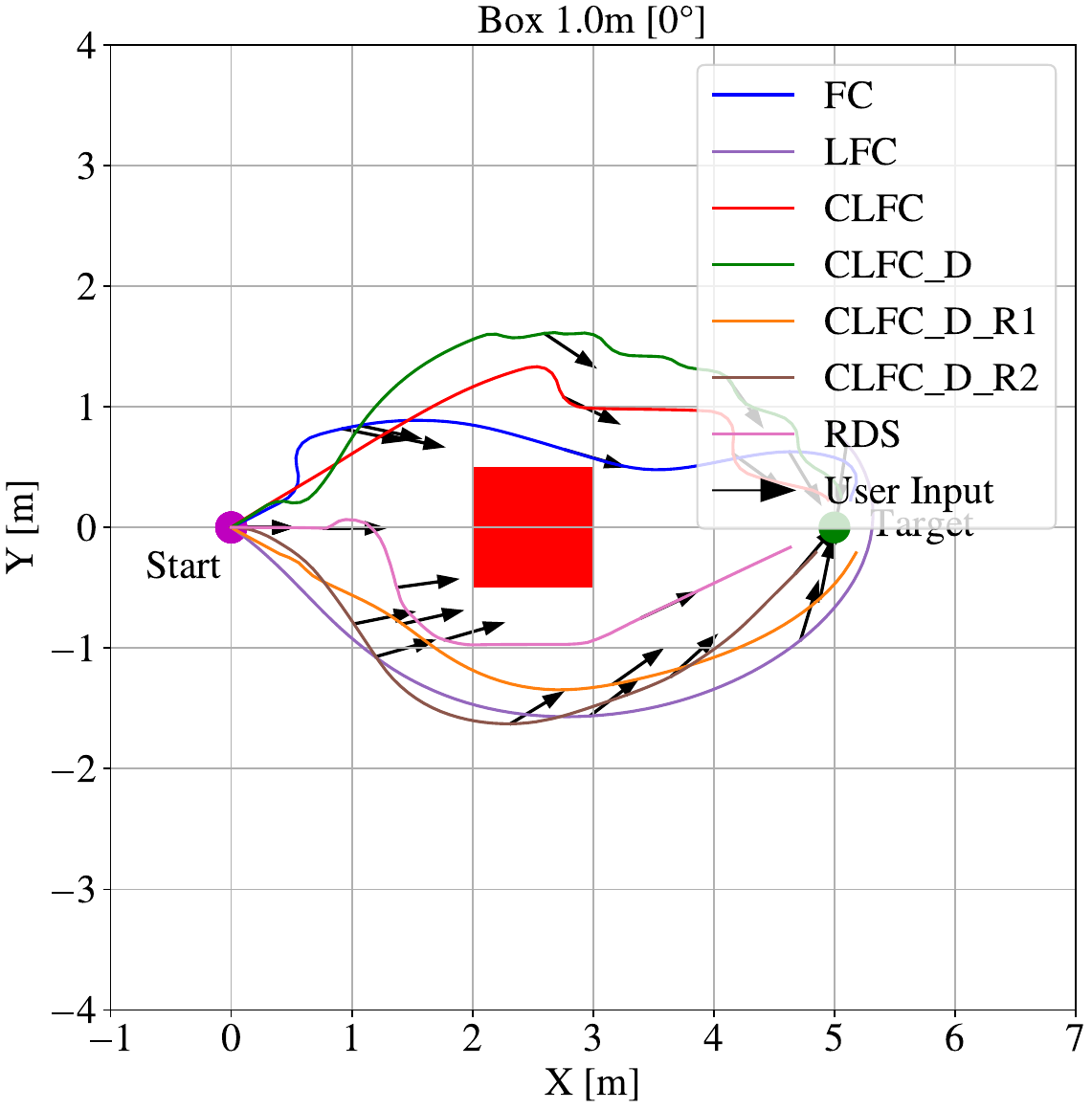}
    \endminipage\hfill
    \minipage{0.32\textwidth}
      \includegraphics[width=\linewidth]{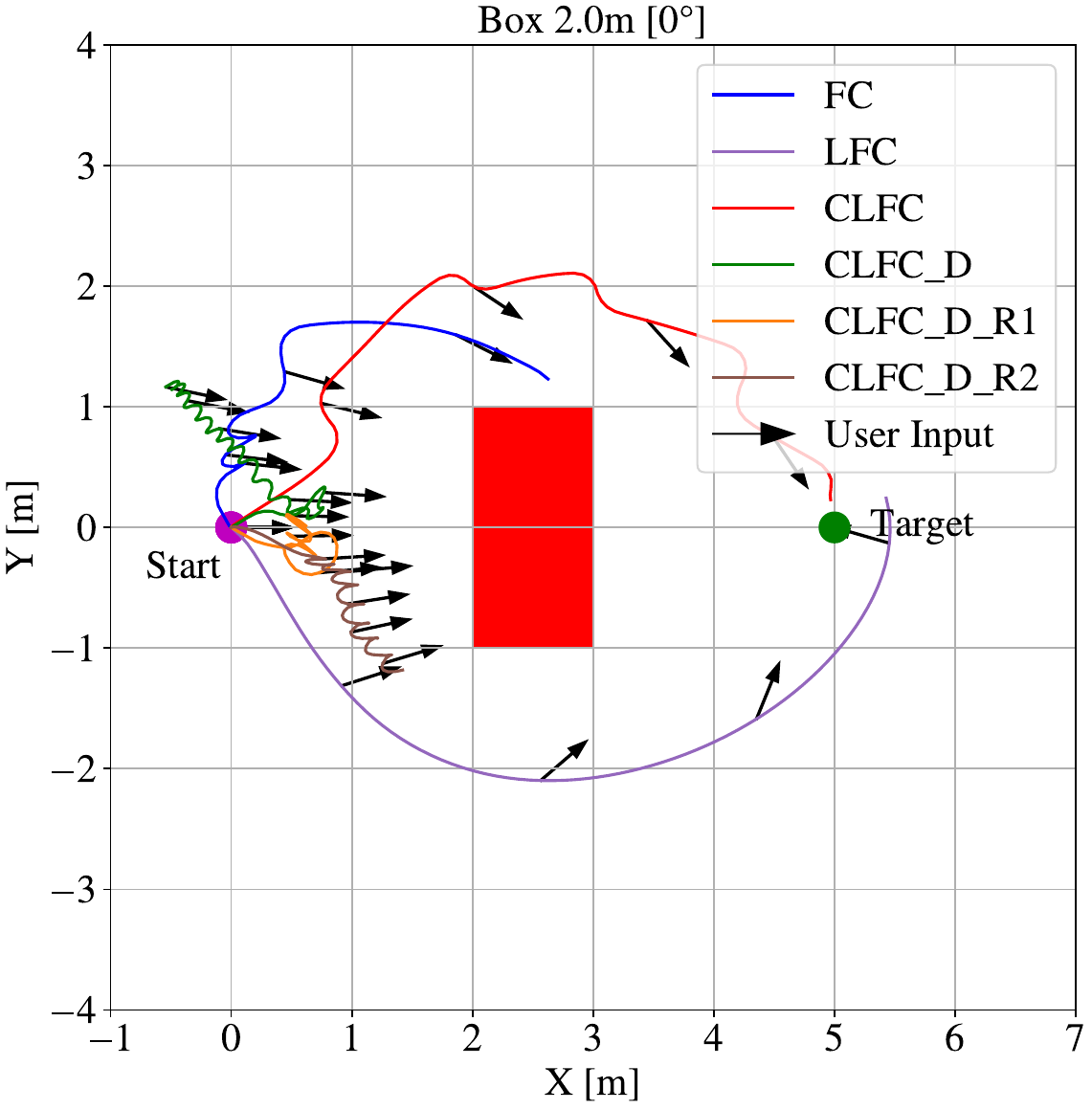}
    \endminipage\hfill
    \minipage{0.32\textwidth}%
      \includegraphics[width=\linewidth]{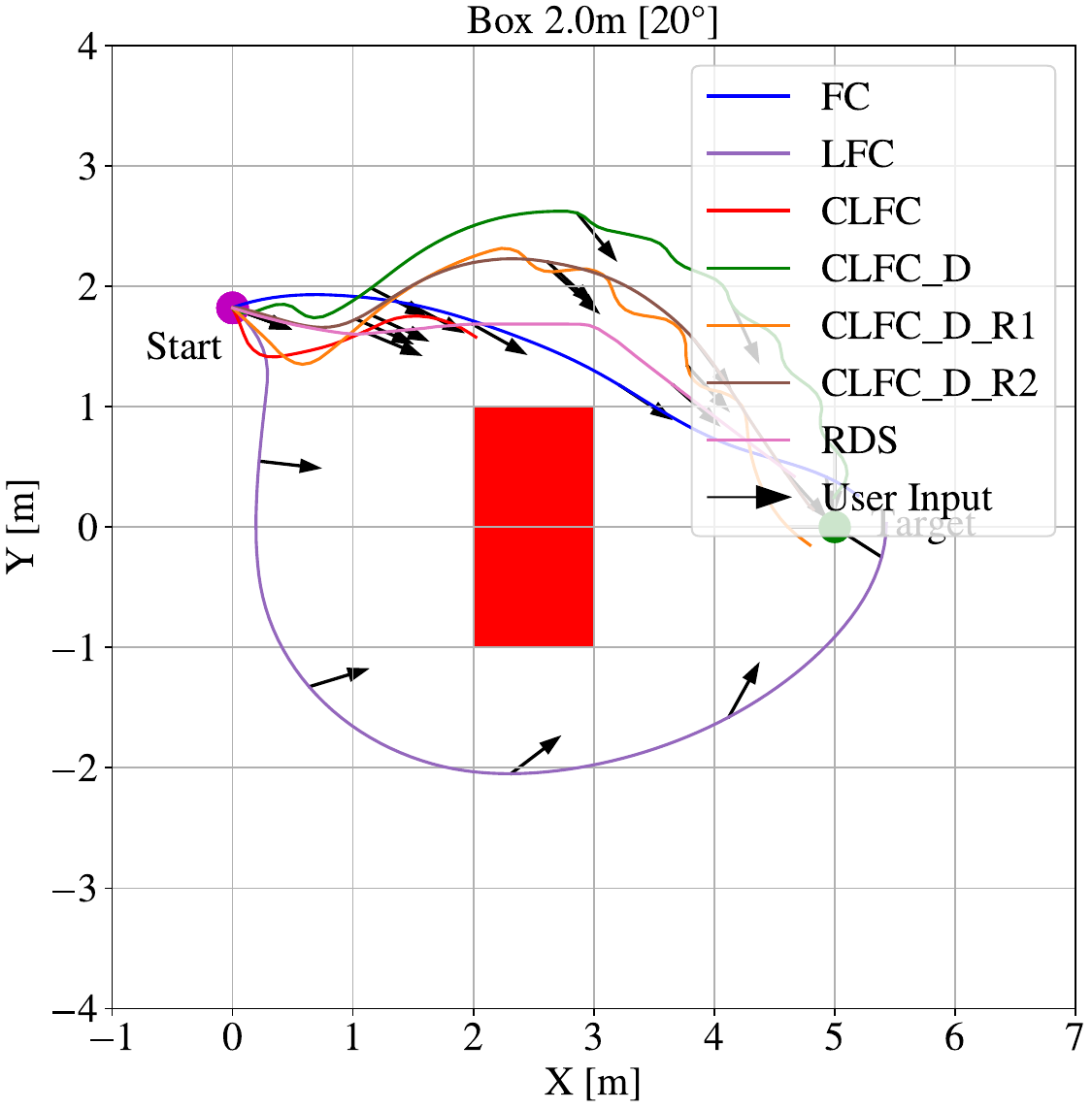}
    \endminipage
    \\
    \minipage{0.32\textwidth}
      \includegraphics[width=\linewidth]{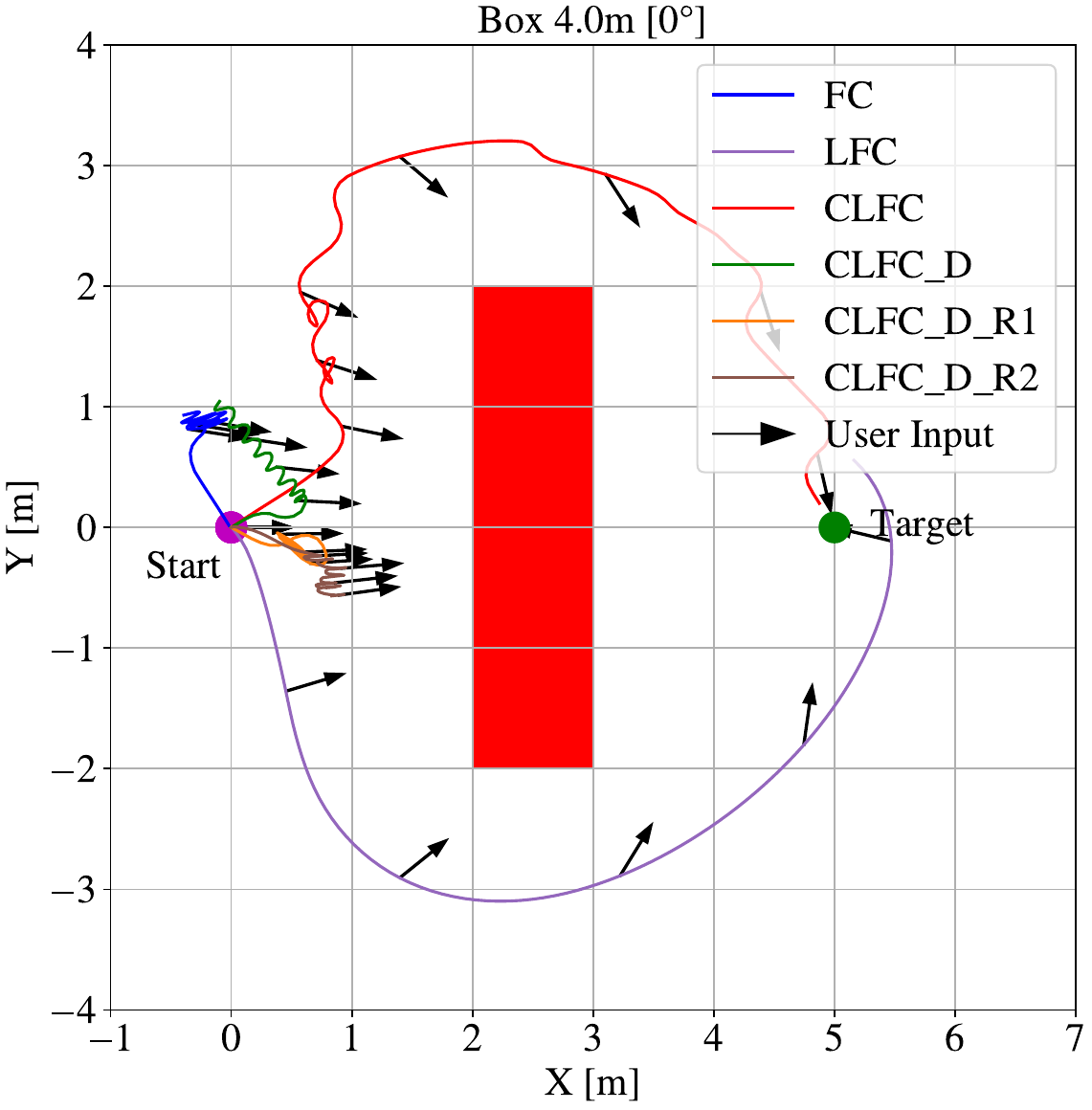}
    \endminipage\hfill
    \minipage{0.32\textwidth}
      \includegraphics[width=\linewidth]{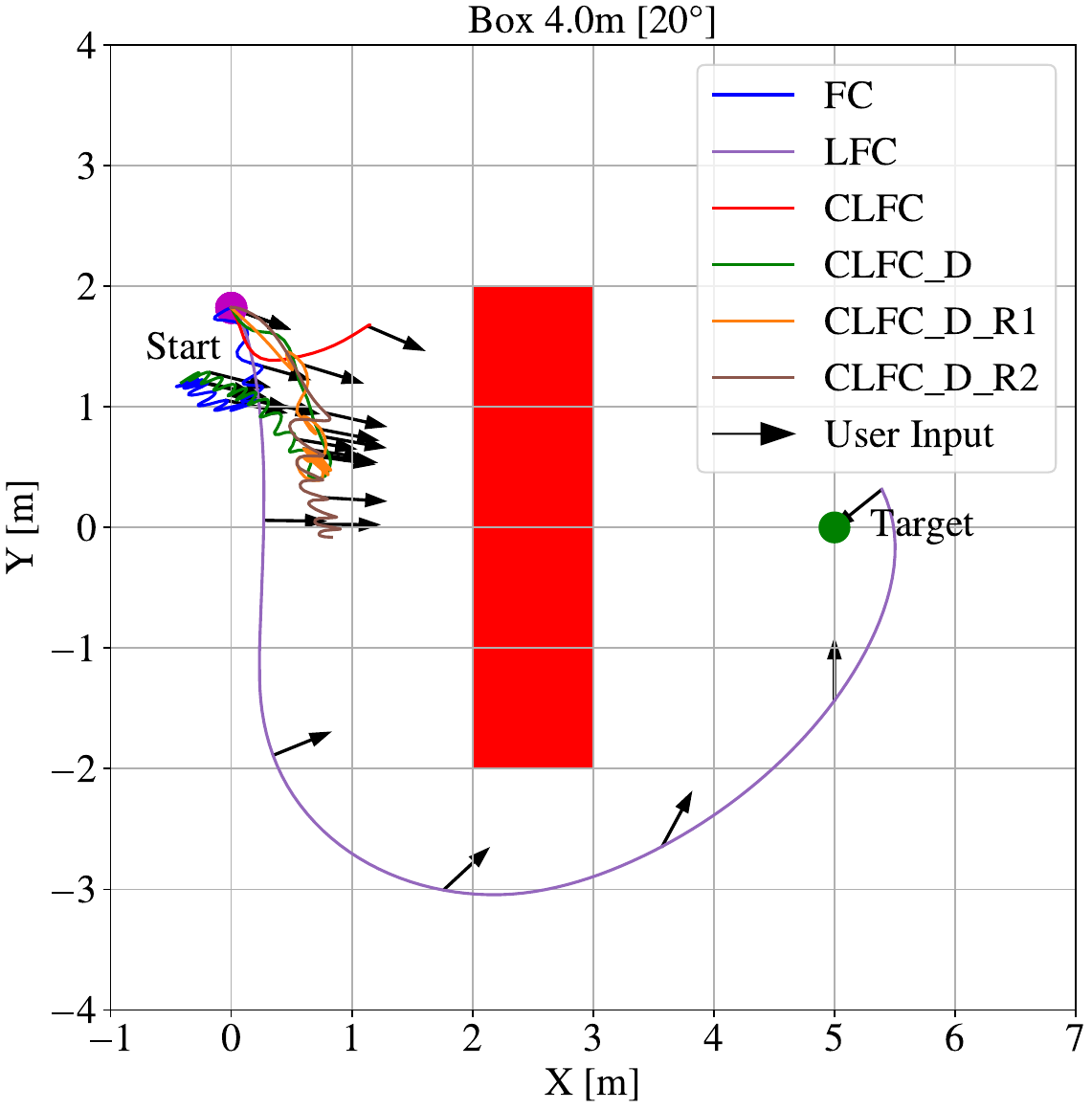}
    \endminipage\hfill
    \minipage{0.32\textwidth}%
      \includegraphics[width=\linewidth]{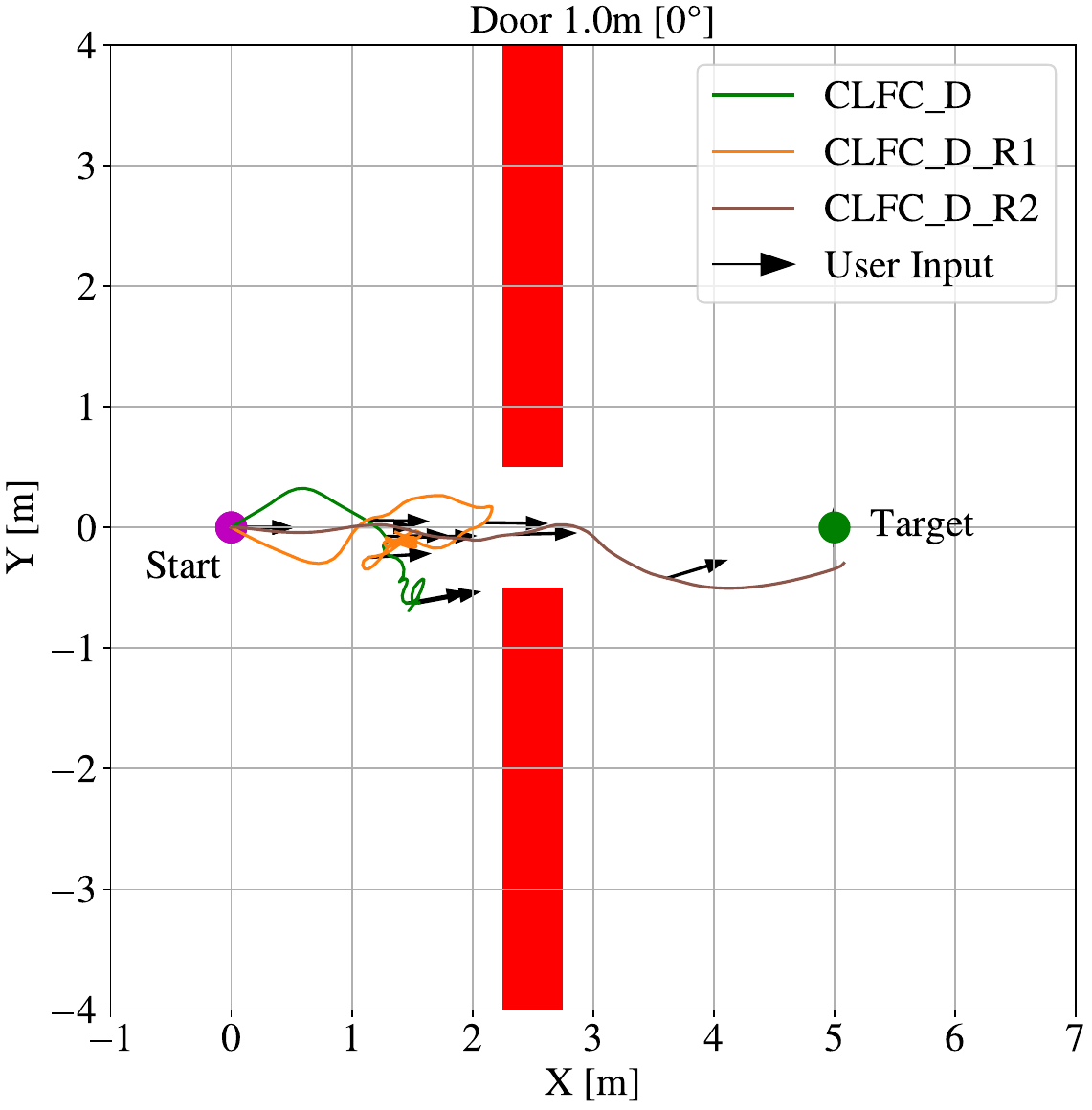}
    \endminipage
    \\
    \minipage{0.32\textwidth}
      \includegraphics[width=\linewidth]{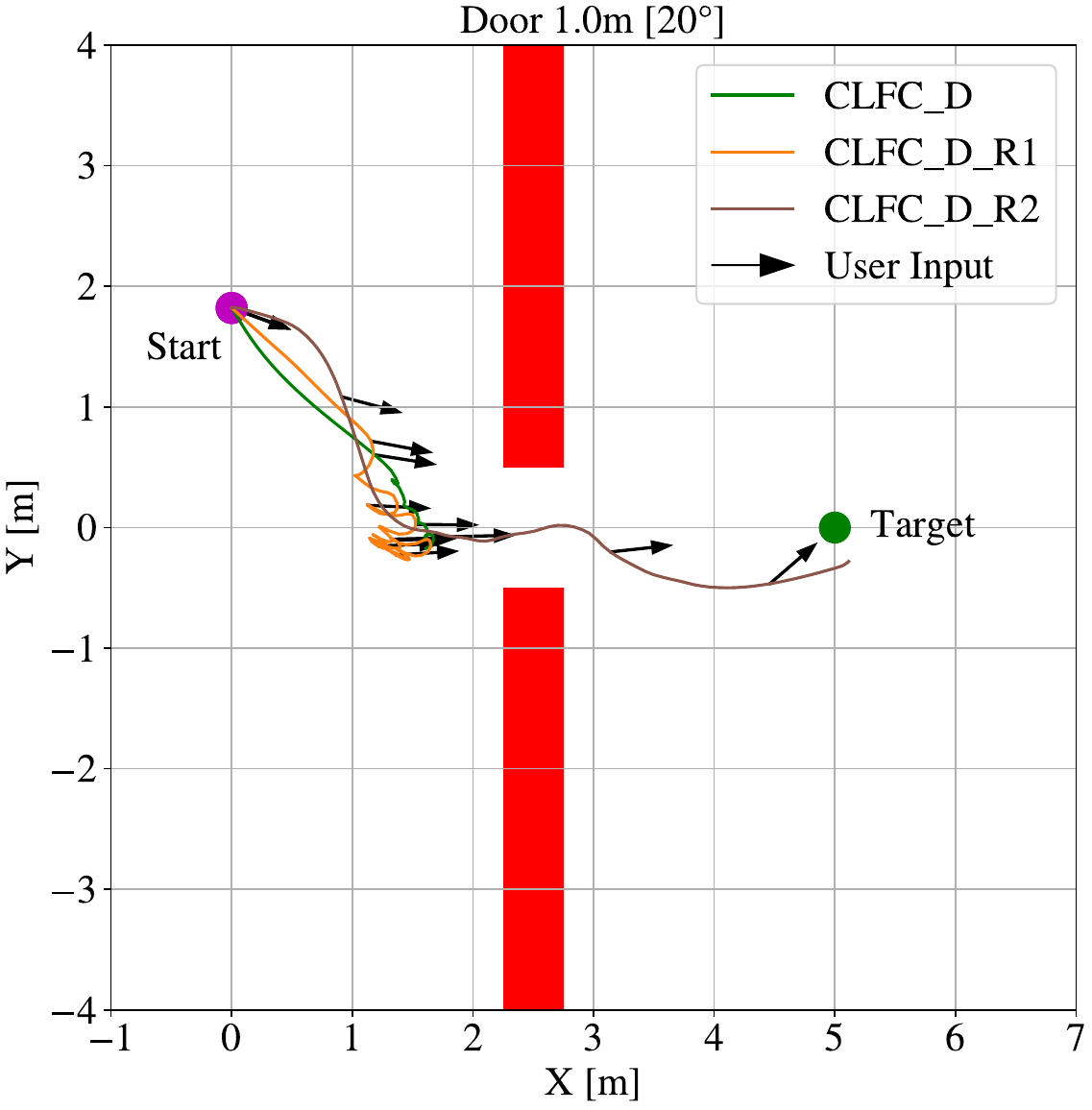}
    \endminipage\hfill
    \minipage{0.32\textwidth}
      \includegraphics[width=\linewidth]{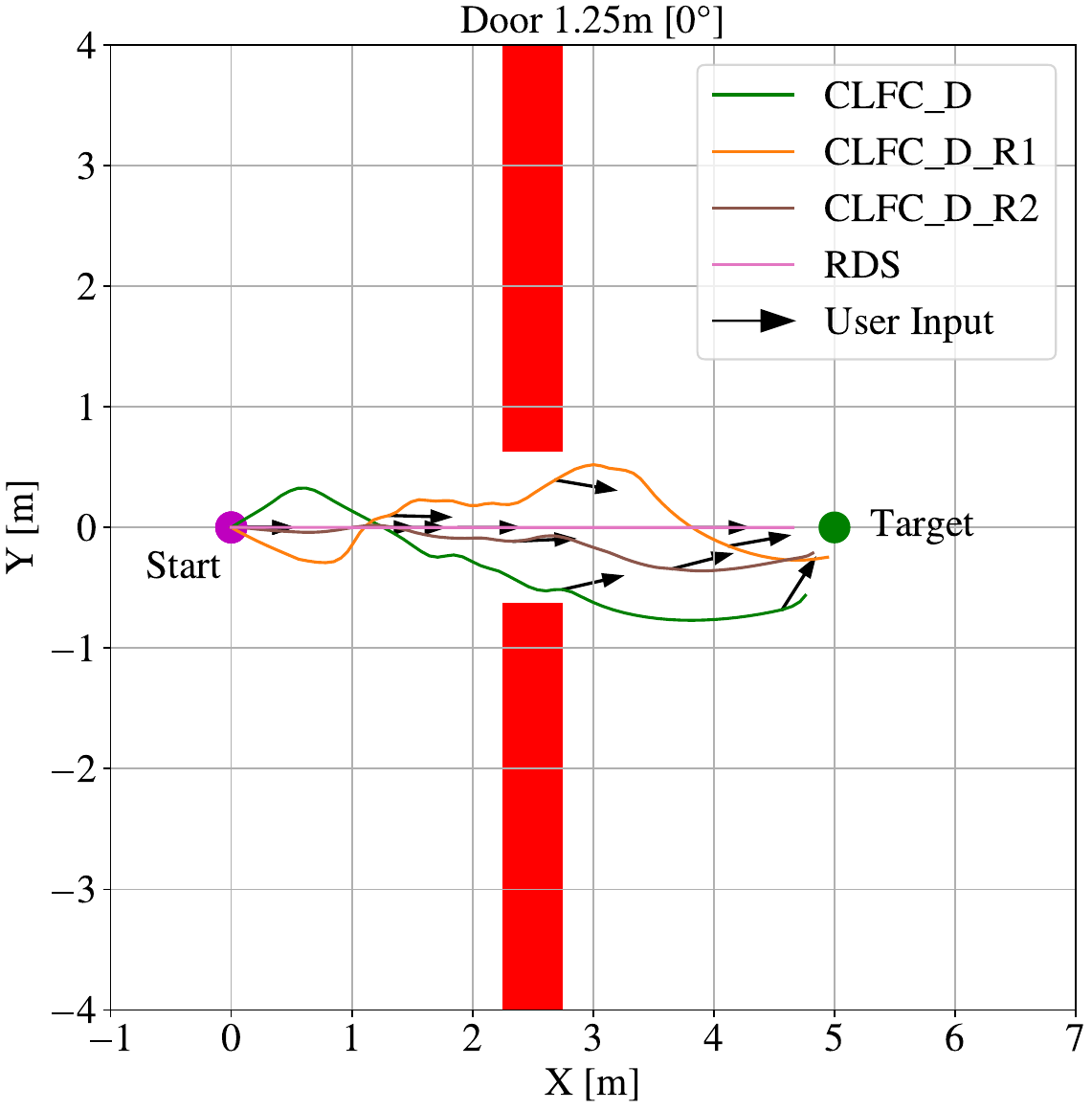}
    \endminipage\hfill
    \minipage{0.32\textwidth}%
      \includegraphics[width=\linewidth]{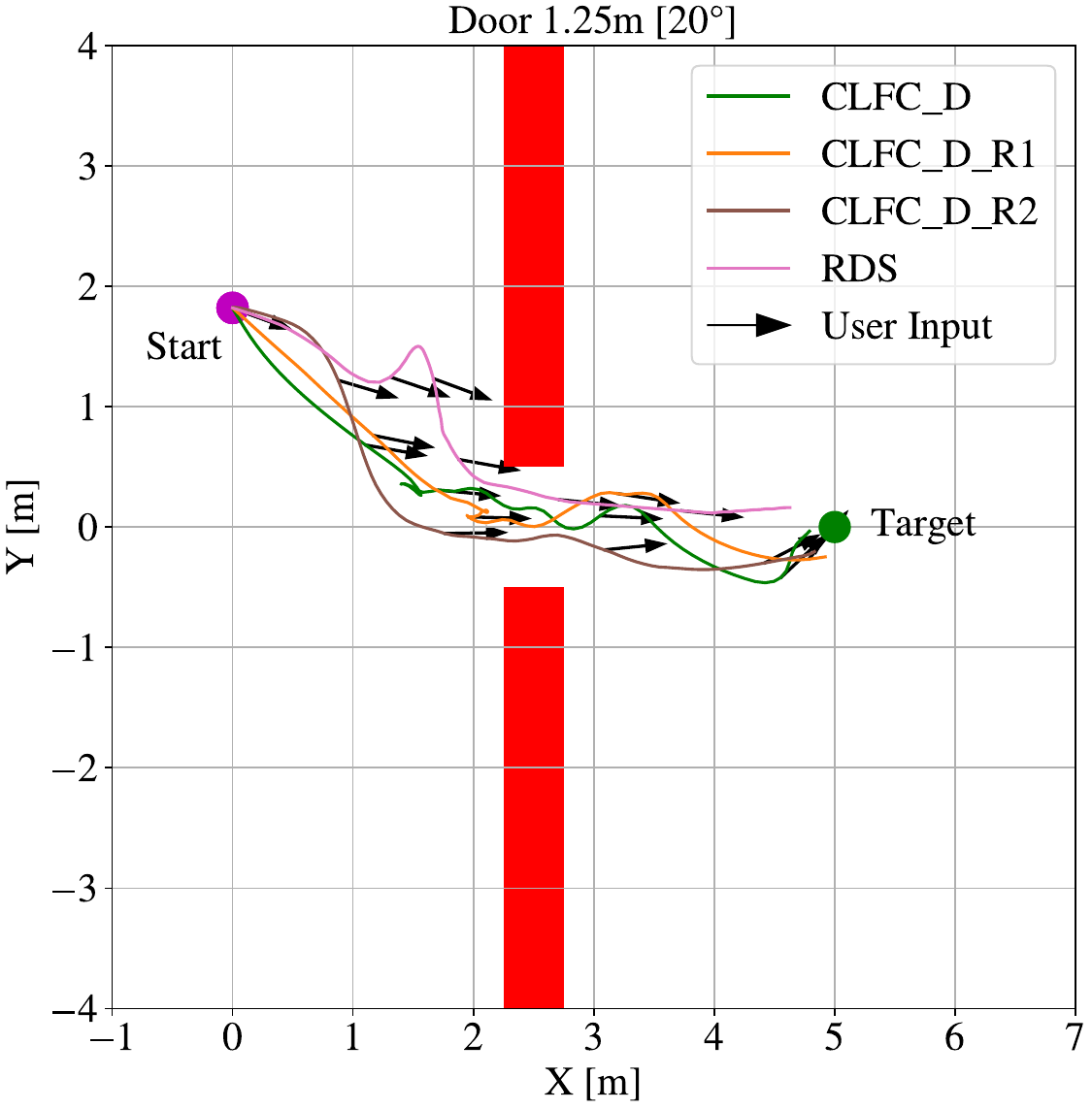}
    \endminipage

\vspace{1em}

\newpage

\subsection{Velocity and acceleration plots of simulated user experiments \ref{subsec:simulated_user_input}.}

\noindent\minipage{0.495\textwidth}
  \includegraphics[width=\linewidth]{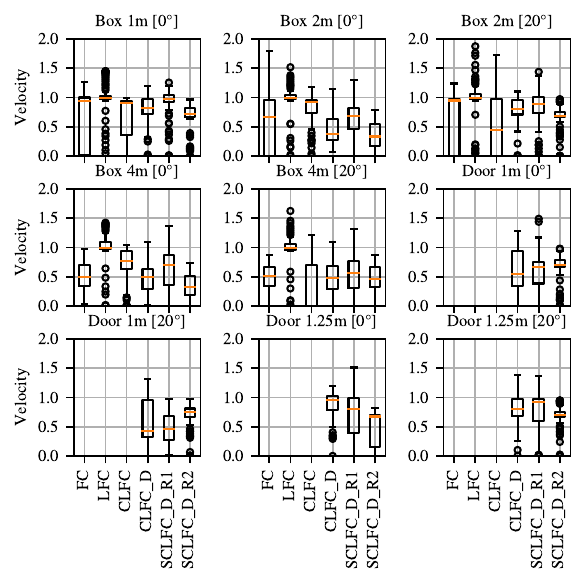}
\endminipage\hfill
\minipage{0.495\textwidth}%
  \includegraphics[width=\linewidth]{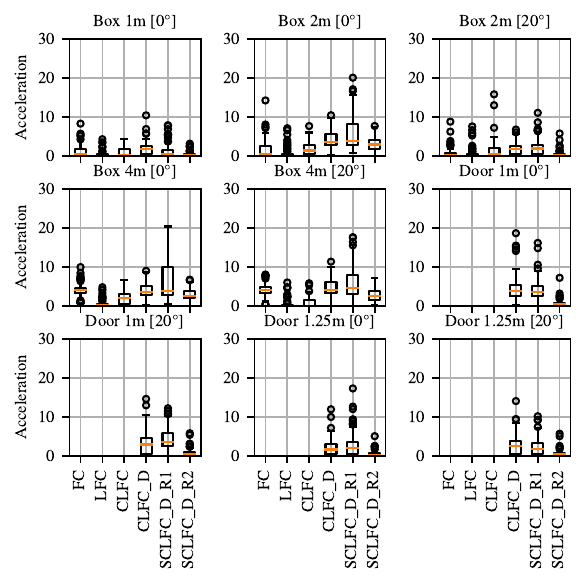}
\endminipage

\subsection{Heading and jerk of manual user input experiments \ref{subsec:manual_user_input}.}

\noindent\minipage{0.495\textwidth}
  \includegraphics[width=\linewidth]{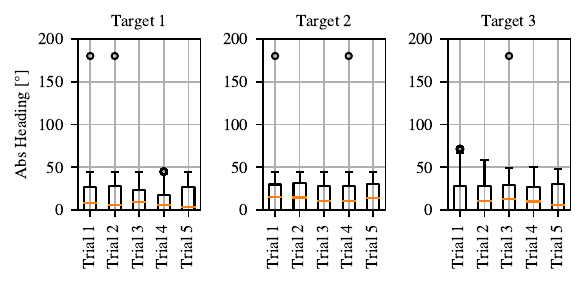}
\endminipage\hfill
\minipage{0.495\textwidth}%
  \includegraphics[width=\linewidth]{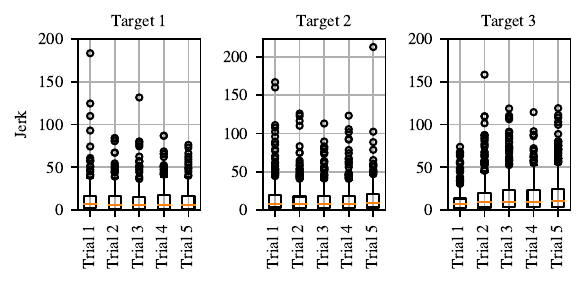}
\endminipage
\\
\minipage{0.495\textwidth}
  \includegraphics[width=\linewidth]{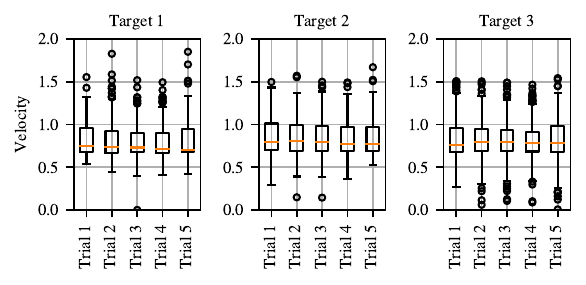}
\endminipage\hfill
\minipage{0.495\textwidth}%
  \includegraphics[width=\linewidth]{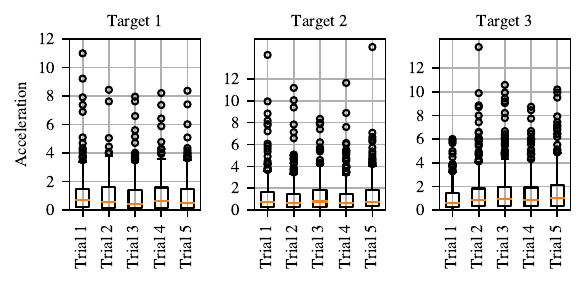}
\endminipage

\vspace{10em}

\subsection{Metrics of real life experiments during which each task was solved 5 times \ref{subsec:real_world_validation}.}

\vspace{1.23em}

\noindent\minipage{0.495\textwidth}
  \includegraphics[width=\linewidth]{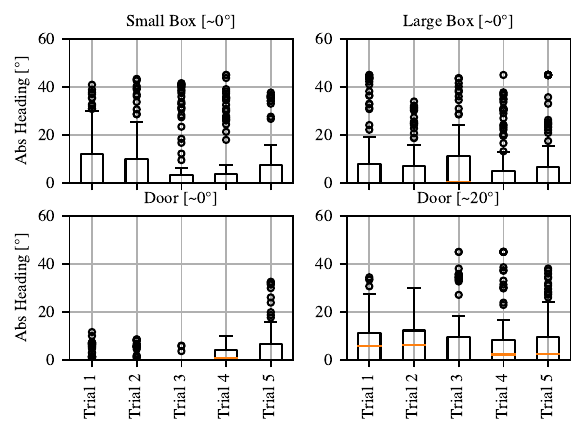}
\endminipage\hfill
\minipage{0.495\textwidth}%
  \includegraphics[width=\linewidth]{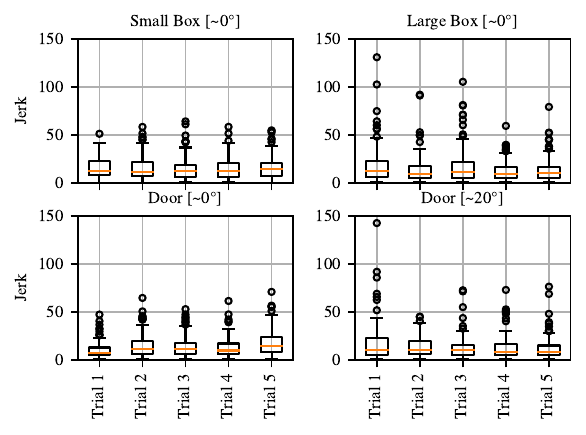}
\endminipage
\\
\minipage{0.495\textwidth}
  \includegraphics[width=\linewidth]{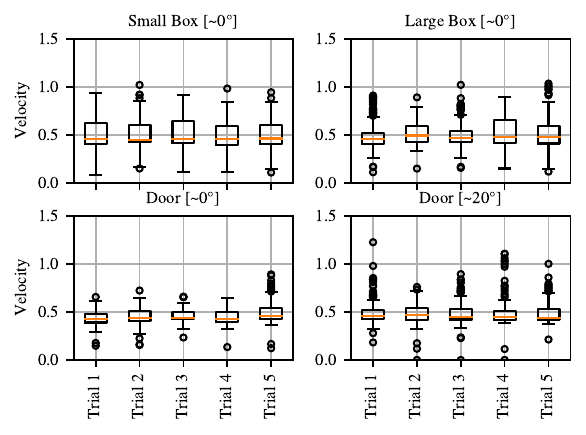}
\endminipage\hfill
\minipage{0.495\textwidth}%
  \includegraphics[width=\linewidth]{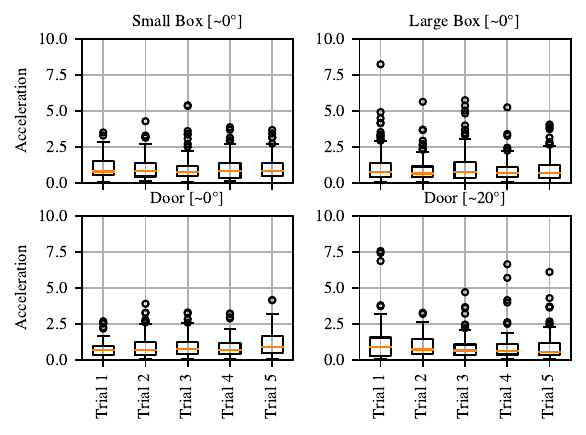}
\endminipage